# OPINION MINING AND ANALYSIS: A SURVEY

Arti Buche[#1], Dr. M. B. Chandak[#2], Akshay Zadgaonkar[#3]

Computer Science Department, RCOEM
India
[1]`artibuche`@gmail.com
[2]`chandakmb@gmail.com`

[*]Zadgaonkar Software Pvt. Ltd. India
[3] `zadgaonkar.akshay@gmail.com`

## ABSTRACT

*The current research is focusing on the area of Opinion Mining also called as sentiment analysis due to sheer volume of opinion rich web resources such as discussion forums, review sites and blogs are available in digital form. One important problem in sentiment analysis of product reviews is to produce summary of opinions based on product features. We have surveyed and analyzed in this paper, various techniques that have been developed for the key tasks of opinion mining. We have provided an overall picture of what is involved in developing a software system for opinion mining on the basis of our survey and analysis.*

## 1. INTRODUCTION

Natural Language Processing (NLP) deals with actual text element processing. The text element is transformed into machine format by NLP. Artificial Intelligence (AI) uses information provided by the NLP and applies a lot of maths to determine whether something is positive or negative. Several methods exist to determine an author's view on a topic from natural language textual information. Some form of machine learning approach is employed and which has varying degree of effectiveness. One of the types of natural language processing is opinion mining which deals with tracking the mood of the people regarding a particular product or topic. This software provides automatic extraction of opinions, emotions and sentiments in text and also tracks attitudes and feelings on the web. People express their views by writing blog posts, comments, reviews and tweets about all sorts of different topics. Tracking products and brands and then determining whether they are viewed positively or negatively can be done using web. The opinion mining has slightly different tasks and many names, e.g. sentiment analysis, opinion extraction, sentiment mining, subjectivity analysis, affect analysis, emotion analysis, review mining, etc. However, they all come under the umbrella of sentiment analysis or opinion mining. Sentiment classification, feature based sentiment classification and opinion summarization are few main fields of research predominate in sentiment analysis.

In recent years, we have witnessed that opinionated postings in social media have helped reshape businesses, and sway public sentiments and emotions, which have profoundly impacted on our social and political systems. Such postings have also mobilized masses for political changes such as those happened in some Arab countries in 2011. It has thus become a necessity to collect and study opinions on the Web. Of course, opinionated documents not only exist on the Web (called external data), many organizations also have their internal data, e.g., customer feedback collected from emails and call centres or results from surveys conducted by the organizations.

Opinion mining can be useful in several ways. For example, in marketing, it tracks and judges the success rate of an ad campaign or launch of new product, determine popularity of products and services with its versions also tell us about demographics which like or dislike particular features. For example, a review might be about a digital camera might be broadly positive, but be specifically negative about how heavy it is. The vendor gets a much clearer picture of public opinion than surveys or focus groups, if this kind of information is indentified in a systematic way.

The technique to detect and extract subjective information in text documents is opinion mining and sentiment analysis. In general, the overall contextual polarity or sentiment of a writer about some aspect can be determined using sentiment analysis. The main challenge in this area is the sentiment classification in which the sentiment may be a judgement, mood or evaluation of an object namely film, book, product, etc which can be in the form of document or sentence or feature that can be labelled as positive or negative.

Classifying entire documents according to the opinions towards certain objects is called as sentiment classification. One form of opinion mining in product reviews is also to produce feature-based summary. To produce a summary on the features, product features are first identified, and positive and negative opinions on them are aggregated. Features are product attributes, components and other aspects of the product. The effective opinion summary, grouping feature expressions which are domain synonyms is critical. It is very time consuming and tedious for human users to group typically hundreds of feature expressions that can be discovered from text for an opinion mining application into feature categories. Some automated assistance is needed. Opinion summarization does not summarize the reviews by selecting a subset or rewrite some of the original sentences from the reviews to capture the main points as the classic text summarization. [2]

The paper is organized into the following sections: the data sources used for opinion mining, introduces machine learning and sentiment analysis tasks for sentiment classification, text classification, tools available for sentiment classification and the performance evaluation. The last section concludes the study and discusses some future directions for research.

## 2. DATA SOURCE

People and companies across disciplines exploit the rich and unique source of data for varied purposes. The major criterion for the improvement of the quality services rendered and enhancement of deliverables are the user opinions. Blogs, review sites and micro blogs provide a good understanding of the reception level of products and services.

### 2.1 Blogs

The name associated to universe of all the blog sites is called blogosphere. People write about the topics they want to share with others on a blog. Blogging is a happening thing because of its ease and simplicity of creating blog posts, its free form and unedited nature. We find a large number of posts on virtually every topic of interest on blogosphere. Sources of opinion in many of the studies related to sentiment analysis, blogs are used. [3]

## 2.2. Review Sites

Opinions are the decision makes for any user in making a purchase. The user generated reviews for products and services are largely available on internet. The sentiment classification uses reviewer's data collected from the websites like www.gsmarena.com (mobile reviews), www.amazon.com (product reviews), www. CNETdownload.com (product reviews), which hosts millions of product reviews by consumers. [1]

## 2.3. Micro-blogging

A very popular communication tool among Internet users is micro-blogging. Millions of messages appear daily in popular web-sites for micro-blogging such as Twitter, Tumblr, Facebook. Twitter messages sometimes express opinions which are used as data source for classifying sentiment. [4]

## 3. SENTIMENT CLASSIFICATION

Sentiment classification or Polarity classification is the binary classification task of labelling an opinionated document as expressing either an overall positive or an overall negative opinion. A technique for analysing subjective information in a large number of texts, and many studies is sentiment classification. A typical approach for sentiment classification is to use machine learning algorithms.

## 3.1. Machine Learning

A system capable of acquiring and integrating the knowledge automatically is referred as machine learning. The systems that learn from analytical observation, training, experience, and other means, results in a system that can exhibit self-improvement, effectiveness and efficiency. Knowledge and a corresponding knowledge organization are usually used by a machine learning system to test the knowledge acquired, interpret and analyse. One of the machine learning algorithms is taxonomy based depending on outcome of the algorithm or type of input available.

- Supervised learning generates a function which maps inputs to desired outputs also called as labels because they are training examples labelled by human experts. Since it is a text classification problem, any supervised learning method can be applied, e.g., Naïve Bayes classification, and support vector machines
- Unsupervised learning models a set of inputs, like clustering, labels are not known during training. Classification is performed using some fixed syntactic patterns which are used to express opinions. The part-of-speech (POS) tags are used to compose syntactic patterns.
- Semi-supervised learning generate an appropriate function or classifier in which both labelled and unlabelled examples are combined. [2, 5]

## 3.2. Sentiment Analysis Tasks

Sentiment analysis tasks mainly consists of classifying the polarity of a given text at the document, sentence or feature/aspect level expressing the opinion as positive, negative or neutral. The sentiment analysis can be performed at one of the three levels: the document level, sentence level, feature level.

- Document Level Sentiment Classification: In document level sentiment analysis main challenge is to extract informative text for inferring sentiment of the whole document. The learning methods can be confused because of objective statements are rendered by subjective statements and complicate further for document categorization task with conflicting sentiment. [6]
- Sentence Level Sentiment Classification: The sentiment classification is a fine-grained level than document level sentiment classification in which polarity of the sentence can be given by three categories as positive, negative and neutral. The challenge faced by sentence level sentiment classification is the identification features indicating whether sentences are on-topic which is kind of co-reference problem [6]
- Feature Level Sentiment Classification: Product features are defined as product attributes or components. Analysis of such features for identifying sentiment of the document is called as feature based sentiment analysis. In this approach positive or negative opinion is identified from the already extracted features. It is a fine grained analysis model among all other models [2]

## 4. TEXT CLASSIFICATION

Now a day's massive volume of online text is available through different websites, internet news feed, emails, cooperate databases and digital library. The main problem is to classify text documents from such massive databases. Using set of training labelled examples statistical text learning algorithms can be trained to approximately classify documents. The news articles and web pages were automatically catalogued by these text classification algorithms.

Naïve Bayes Classifier: The Naïve Bayes—well known probabilistic classifier—and describes its application to text. In order to incorporate unlabelled data, the foundation Naïve Bayes was build. The task of learning of a generative model is to estimate the parameters using labelled training data only. The estimated parameters are used by the algorithm to classify new documents by calculating which class the generated the given document belongs to. The naive Bayesian classifier works as follows:

- Consider a training set of samples, each with the class labels T. There are k classes, $C_1, C_2, \ldots, C_k$. Every sample consists of an n-dimensional vector, $X = \{x_1, x_2, \ldots, x_n\}$, representing n measured values of the n attributes, $A_1, A_2, \ldots, A_n$, respectively.
- The classifier will classify the given sample X such that it belongs to the class having the highest posterior probability. That is X is predicted to belong to the class $C_i$ if and only $P(C_i|X) > P(C_j|X)$ for $1 \leq j \leq m, j \neq i$.

Thus we find the class that maximizes $P(C_i|X)$. The maximized value of $P(C_i|X)$ for class $C_i$ is called the maximum posterior hypothesis. By Bayes' theorem

$$P(C_i|X) = \frac{P(X|C_i) P(C_i)}{P(X)}$$

- Only $P(X|C_i) P(C_i)$ value needs to be maximized as for all classes value of $P(X)$ is same. If the priori probabilities, $P(C_i)$ of the class are not known, then it is assumed that the classes are likely to be equal, that is, $P(C_1) = P(C_2) = \ldots = P(C_k)$, and we would therefore maximize $P(X|C_i)$. Otherwise the value of $P(X|C_i) P(C_i)$ is maximized. The priori probabilities of a class are estimated by

$$P(C_i) = \text{freq}(C_i, T)/|T|.$$

- To compute $P(X|C_i)$ would be computationally expensive as given data sets consist of many attributes. To reduce computation in evaluation of $P(X|C_i) P(C_i)$, the conditional class independence of naive assumption is made. The values of the attributes of the class label of the given sample presume to be conditionally independent of one another. Mathematically this means that

$$P(X|C_i) \approx \prod_{k=1}^{n} P(x_k|C_i)$$

Estimation of the probabilities $P(x_1|C_i), P(x_2|C_i)\ldots P(x_n|C_i)$ can easily be done from the training set. For sample X, xk refers to the value of attribute $A_k$.

(a) If $A_k$ is categorical, then the number of samples $P(x_k|C_i)$ of class $C_i$ in T have the value $x_k$ for attribute $A_k$, divided by the number of sample of class Ci, freq $(C_i, T)$, in T.
(b) We typically assume that if $A_k$ is continuous-valued then the values have a Gaussian distribution with a mean µ and standard deviation $\sigma$ defined by

$$g(x, \mu, \sigma) = \frac{1}{\sqrt{2\pi\sigma}} \exp -\frac{(x-\mu)^2}{2\sigma^2}$$

so that $p(x_k|C_i) = g(x_k, \mu C_i, C_i)$.

We need to compute µ $C_i$ and $C_i$, which are the mean and standard deviation of values of attribute $x_k$ for training samples of class $C_i$.

- To predict the class label X, the evaluation of $P(X|Ci) P(Ci)$ is done for each class Ci. The class that maximizes the value of $P(X|Ci) P(Ci)$ is the class label of X is $C_i$ predicted by the classifier. [7]
- Expectation Maximization: Due to high variance in the parameter estimates of the generative model, the Naive Bayes method's accuracy will suffer because it has small set of labelled training data. This problem can be overcome by augmenting this labelled training data with a large set of unlabelled data and combining the two sets with EM, we can improve the parameter estimates. EM is used for maximum likelihood or maximum posterior estimation in problems with incomplete data and is class of iterative algorithms.

Biological data is modelled using probabilistic models, such as HMM (Hidden Markov Model) or Bayesian networks which are efficient and robust procedures for learning parameters from observations. There are various sources for missing values such as in medical diagnosis, missing data for certain tests or gene expression clustering due to intentional omission of gene-to-cluster assignments in the probabilistic model. Such error does not occur in EM algorithm.

Consider an example of a simple coin-flip-ping experiment in which we are given a pair of unknown biases $_A$ and $_B$ of coins A and B respectively that is, on any given flip, coin A will land on tails with probability 1– $_A$ and heads with probability $_A$ and similarly for coin B. The goal is to estimate = ( $_A$, $_B$) by repeating the following procedure five times: with equal probability, choose one of the two coins randomly and perform with the selected coin ten independent tosses. Thus, a total of 50 coin tosses are performed. During the experiment, suppose the track of two vectors $x = (x_1, x_2, \ldots, x_5)$ and $z = (z_1, z_2, \ldots, z_5)$ is kept, during the $i^{th}$ set of tosses where the identity of the coin used is $z_i \in \{A,B\}$, where $xi \in \{0,1,\ldots,10\}$ is the number of heads

43

observed. In the model the parameter estimation setting also known as complete data case in which the values of all relevant random variables which consists of each coin flip and type of coin used for flip are known. Here, a simple way to estimate $\theta_A$ and $\theta_B$ is to return the observed proportions of heads for each coin:

$$\hat{\theta}_A = \frac{\text{\# of heads using coin A}}{\text{total \# of flips using coin A}}$$

and

$$\hat{\theta}_B = \frac{\text{\# of heads using coin B}}{\text{total \# of flips using coin B}}$$

It is known as maximum likelihood estimation which assesses the quality based on the probability assigned to the observed data of statistical model. The formulae are solved for the parameters $\hat{\theta} = (\hat{\theta}_A, \hat{\theta}_B)$, that maximize If $\log P(x, z; )$ is the logarithm of the joint probability obtained for any particular vector of observed head counts *x* and coin types *z*.

Now consider the parameter estimation problem in which the recorded head counts *x* are given and z as hidden variables or latent factors is not given. Such parameter estimation setting is known as the incomplete data case. The coin used for each set of tosses is unknown, thus the computation of heads for each coin is no longer possible. This problem, the parameter estimation can be reduced with incomplete data to maximum likelihood estimation with complete data, if there was some way of guessing the correctly used coin from each of five sets. One of the iterative schemes used to obtain completions could work as follows: start from some initial parameters, $\hat{\theta}^{(t)} = \left(\hat{\theta}_A^{(t)}, \hat{\theta}_B^{(t)}\right)$, determine whether coin *A* or coin *B* from each five sets was more likely to have generated the observed flips. Then, assume the guessed coin to be correct, and apply the estimation procedure of regular maximum likelihood to get $\hat{\theta}^{(t+1)}$. Finally, repeat these two steps until convergence. As the quality of the resulting completions improves so the estimated model also improves.

The expectation maximization algorithm oscillates between the steps of guessing the model depending on probability distribution over completions of missing data (known as the E-step) and then re-estimating the model parameters using these completions (known as the M-step). The name 'E-step' needs only to compute expected statistics over completions rather than explicitly forming probability distribution over completions. Similarly, the name 'M-step' consists of model re-estimation which can be thought of as 'maximization' of the expected log-likelihood of the data. [8, 9]

## 5. GROUPING FEATURES

Regular expressions: Regular expressions are highly specialized programming language, in which the rules are specified for the set of possible strings that can be matched and the set might contain English sentences, or e-mail addresses etc. String processing tasks which are performed using regular expressions becomes very complicated because regular expressions language is relatively small and restricted.

This mechanism was motivated to avoid silly errors by automated systems, particularly machine learning models. For example, opening salutations, such as "Dear Jane", were falsely assigned with instructions by a machine learning model in some runs of cross-validation tests, possibly due to the frequent occurrences of person names in instructions. With this mechanism, "dear __NAME__", a normalized form of "Dear Jane", was compared against all such normalized instances in the training data, and false assignment of instructions could be avoided after reviewing emotions assigned to the found training instances, specifically, by confirming more than two-thirds of the found training instances were not assigned with instructions. [11]

Clustering: Clustering is the natural technique used to discover hundreds of feature expressions from text for an opinion mining application. Similarity measures used for clustering are usually based on some form of distributional similarity. There are two main kinds of similarity measures those relying on pre-existing knowledge resources (e.g., thesaurus, and semantic networks) and those relying on distributional properties of words in corpora.

First, a pre-processing pass could build a list of words and phrases that appear frequently in the review of a particular restaurant but are uncommon in the wider corpus. This should find phrases like the name of a dish that many people are talking about. Second, given the narrow domain of the problem, it should also be possible to hand-build a list of common ideas a reader might want to know about, like service, food, and price. Extracting these combined, specific features should lead to purpose-built vectors that form clusters around relevant concepts. [12]

## 6. EVALUATION MEASURES

All since the problem of grouping feature expressions is a clustering task, two common measures for evaluating clustering are used the study, Entropy and Purity. Below, we briefly describe entropy and purity. Given a data set DS, its gold partition is G = {$g_1$....,…,$g_j$....,…$g_k$}, where k is the given number of clusters. The groups partition DS into k disjoint subsets, $DS_1$,…, $DS_i$, …, $DS_k$.

- Entropy: For each resulting cluster, we can measure its entropy using Equation (a), where $P_i(g_i)$ is the proportion of $g_i$ data points in $DS_i$. The total entropy of the whole clustering (which considers all clusters) is calculated by Equation (b)
- Purity: Purity measures the extent that a cluster contains only data from one gold-partition. The cluster purity is computed with Equation (c). The total purity of the whole clustering (all clusters) is computed with Equation (d) [2]

$$entropy(DS_i) = -\sum_{j=1}^{k} P_i(g_i) log_2 P_i(g_i) \quad \text{(a)}$$

$$entropy_{total} = \sum_{i=1}^{k} \frac{|DS_i|}{|DS|} entropy(DS_i) \quad \text{(b)}$$

$$purity(DS_i) = \max_{j} P_i(g_i) \quad \text{(c)}$$

$$purity_{total} = \sum_{i=1}^{k} \frac{|DS_i|}{|DS|} purity(DS_i) \quad \text{(d)}$$

Similarly for the evaluation of sentiment classification using regular expression is usually measured by precision and recall. Precision is the fraction of relevant retrieved instances, while recall is the fraction of retrieved relevant instances. Therefore precision and recall are based on an understanding and measure of relevance.

$$Precision = \frac{TP}{TP + FP}$$

$$Recall = \frac{TP}{TP + FN}$$

$$F-score = \frac{2 * (precision * recall)}{(precision + recall)}$$

Where TP - number of true positives
      TN - number of true negatives
      FP - number of false positives
      FN - number of false negatives. [11]

## 7. TOOLS

A Red Opal is a tool that enables users to find products based on features. The features from customer reviews are used for scoring each product. Opinions on web are analysed and compared using Opinion observer. The product opinions are displayed feature by feature in graph format.
Automation of aggregation sites is done by Review Seer tool. The extracted features are assigned score by Naïve Bayes classifier as positive and negative review. The crawled pages are not classified properly by this tool. Result is displayed in the form of attribute and its score.

Product features are extracted in Web Fountain using beginning definite Base Noun Phrase (bBNP) heuristic. The sentiment lexicon and sentiment pattern database are used to assign sentiments to feature. Sentiment extraction patterns are defined in sentiment pattern database and polarity of terms is defined in sentiment lexicon. [1]

## 8. CONCLUSIONS

The challenge in sentiment analysis of product reviews is to produce a summary of opinions based on product features/attributes (also called aspects). People can express features with many different words or phrases. These words and phrases are grouped under the same feature group to produce useful summary. Limited work has been done on grouping of synonym features and clustering.

In future, Opinion Mining can be carried out on a set of reviews and set of discovered feature expressions extracted from reviews. The state-of-art for current methods, useful for producing better summary based on feature based opinions as positive, negative or neutral is the Expectation Maximization algorithm based on Naïve Bayesian is the most efficient method. The efficiency of EM algorithm can be increased by augmenting it, to reassign classes of the labelled set.

The natural language text can be processed based on machine learning toolkit called as OpenNLP library. The NLP tasks, such as tokenization, part-of-speech (POS) tagging, named entity extraction, parsing, chunking, sentence segmentation, and co reference resolution are provided by Open NLP library. The advanced text processing services are built using these tasks. OpenNLP also includes perceptron and maximum entropy based machine learning.

After POS tagging, opinion retrieval can be performed by extracting product candidate feature, related opinion and producing opinion feature pairs. The keywords extracted from Opinion Retrieval Module can be used to perform similarity check with the database dictionary. The similarity check can use semi supervised learning.

**Authors**

Arti Buche, Dr. M. B. Chandak and Akshay Zadgaonkar.Arti Buche is pursuing Masters in Technology in Computer Science and Engineering from Shri. Ramdeobaba College of Engineering and Management, Nagpur.

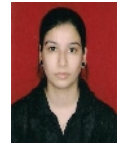